\documentclass[conference]{IEEEtran}
\IEEEoverridecommandlockouts

\usepackage{setspace}
\usepackage{cite}
\usepackage{amsmath,amssymb,amsfonts}
\usepackage{algorithmic}
\usepackage{graphicx}
\usepackage{textcomp}
\usepackage{xcolor}
\usepackage{tabularx}
\usepackage{booktabs} 
\usepackage{array}
\usepackage[font=small,labelfont=bf,labelsep=period]{caption}
\usepackage{hyperref}
\usepackage{enumitem}
\usepackage{cuted}
\usepackage{multirow}
\usepackage{longtable}
\usepackage[margin=1in]{geometry}
\usepackage{float}
\usepackage{stfloats}
\usepackage{placeins}

\begin{document}
\onehalfspacing 

\makeatletter
\def\@maketitle{%
  \newpage
  \null
  \vskip 1em%
  \begin{center}%
    {\fontsize{17}{16}\selectfont\bfseries \@title \par}%
    \vskip 1em%
    {\normalsize
     \lineskip .5em%
     \begin{tabular}[t]{c}%
       \@author
     \end{tabular}\par}%
  \end{center}%
  \par
  \vskip 1.5em}
\makeatother
\title{AQUA: A Large Language Model for Aquaculture \& Fisheries}

\author{
    Praneeth Narisetty, Uday Kumar Reddy Kattamanchi, Lohit Akshant Nimma,\\
    Sri Ram Kaushik Karnati, Shiva Nagendra Babu Kore, Mounika Golamari, Tejashree Nageshreddy\\
\textbf{Kurma AI}\\
\texttt{\{praneeth, uday, lohit, ram, shiva, mounika, tejashree\}@kurma.ai}
}

\maketitle

{\normalsize
\makeatletter
\renewcommand\IEEEkeywordsname{Keywords}
\renewenvironment{abstract}{%
    \centerline{\bfseries \abstractname}%
    \vspace{1em}%
    \itshape\normalsize%
}{}
\makeatother
\begin{abstract}
\textbf{Aquaculture} plays a vital role in global food security and coastal economies by providing sustainable protein sources. As the industry expands to meet rising demand, it faces growing challenges such as disease outbreaks, inefficient feeding practices, rising labor costs, logistical inefficiencies, and critical hatchery issues, including high mortality rates and poor water quality control. Although artificial intelligence has made significant progress, existing machine learning methods fall short of addressing the domain-specific complexities of aquaculture. 

To bridge this gap, we introduce \textbf{AQUA}, the first large language model (LLM) tailored for aquaculture, designed to support farmers, researchers, and industry practitioners. Central to this effort is \textbf{AQUADAPT} (Data Acquisition, Processing and Tuning), an Agentic Framework for generating and refining high-quality synthetic data using a combination of expert knowledge, large-scale language models, and automated evaluation techniques. Our work lays the foundation for LLM-driven innovations in aquaculture research, advisory systems, and decision-making tools. 
\end{abstract}
}
\makeatletter
\renewcommand\IEEEkeywordsname{Keywords}
\renewenvironment{IEEEkeywords}{%
  \section*{\bfseries\IEEEkeywordsname}%
  \itshape\normalsize%
}{}
\makeatother


\section{\textbf{Introduction}}

Aquaculture produces nearly 50\% of the imported seafood consumed in the United States, highlighting the sector’s significant role in the nation’s food supply (NOAA Fisheries, 2023). As the U.S. imports approximately 75\% of its seafood, aquaculture has become increasingly vital to meet demand (USDA Economic Research Service, 2022). However, the domestic aquaculture industry remains underdeveloped compared to its global counterparts, facing persistent structural and operational drawbacks that hinder its scalability and efficiency (Knapp \& Rubino, 2016; Engle \& Stone, 2013).

In the U.S., high operational costs, particularly labor expenses, which account for 40-50\% of total production costs, pose a major barrier to growth (Engle, Hanson, \& Hinshaw, 2021). These costs are exacerbated by limited automation, labor shortages, and regulatory complexity (Knapp, Anderson, \& Tyler, 2022). Disease outbreaks continue to cause substantial economic losses (Mardones, Perez, \& Rojas, 2018), whereas suboptimal feeding strategies lead to wasted resources and environmental degradation (Torrissen et al., 2011). Hatchery management presents additional difficulties, including high mortality rates and challenges in maintaining consistent water quality (Akinbowale, Peng, \& Barton, 2007; Håstein, Lillehaug, \& Jarp, 2005). These issues not only limit productivity but also impact the reliability and sustainability of the domestic supply chain (Engle \& Stone, 2013; Knapp, Anderson, \& Tyler, 2022).

Despite advancements in artificial intelligence (AI) and large language models (LLMs) in fields such as medicine and earth sciences (Moor et al., 2023; Deng et al., 2023), aquaculture has yet to fully benefit from such innovations. Current machine learning applications in the industry are often narrow in scope and lack the contextual understanding required to address multifaceted and domain-specific challenges (Bi et al., 2023; Kumar et al., 2022). 

To bridge this gap, we present \textbf{AQUA}, the first large language model that is purpose-built for the aquaculture domain. AQUA is designed to assist a broad spectrum of users, including farmers, hatchery managers, researchers, and policymakers, by providing intelligent insights, improving decision-making, and enhancing operational efficiency. At the heart of AQUA’s development lies AQUADAPT (Aquaculture Data Acquisition, Processing, and Tuning), a structured and agentic framework for building high-quality domain-relevant datasets. AQUADAPT combines automated web-scale data extraction with rule-based filtering and expert-guided evaluations. It leverages few-shot prompting and large-scale language models to generate rich synthetic question–answer datasets that are iteratively refined through both human-in-the-loop review and automated scoring techniques.

Unlike traditional pipelines, AQUADAPT is designed as an agentic system, a collaborative architecture composed of specialized, semi-autonomous agents, each responsible for distinct stages of the data lifecycle. Agents such as \textbf{Data Agent}, \textbf{Expert Agent}, \textbf{Prompt Agent}, \textbf{QA Agent}, and \textbf{Scoring Agent} work in coordination to collect domain-specific content, curate prompts, generate and refine QAs, evaluate their quality, and guide finetuning. This modularity allows AQUADAPT to be flexible, scalable, and self-improving, enabling dynamic adaptation based on data gaps, model feedback, or expert intervention.

This work is driven by the urgent need for intelligent, scalable, and cost-effective AI solutions in aquaculture, particularly in high-cost regions such as the United States, the UK, and Europe, where labor expenses, regulatory barriers, and operational inefficiencies limit growth. By structuring the data generation pipeline as an agentic framework, AQUADAPT enables the development of rapid, traceable, and high-quality models. We have strong expertise in AI and data science, complemented by direct collaboration with field experts. Through AQUA and the AQUADAPT framework, we aim to catalyze a new wave of domain-adapted, agent-driven LLMs that transform aquaculture research, marine research, fisheries, advisory, and operational decision-making. 

\section{\textbf{Related Work}}

The emergence of domain-specific large language models (LLMs) has accelerated progress in specialized fields by addressing the limitations of general-purpose models. 

In the biomedical domain, BioGPT is a transformer-based language model pre-trained on large-scale biomedical literature data. It leverages domain-specific vocabulary and structure to enable accurate text generation, relation extraction, question answering and document classification. Through fine-tuning on datasets such as PubMedQA, BC5CDR, and HoC, BioGPT achieves superior performance on biomedical NLP tasks compared to general-purpose models, making it a valuable tool for clinical text mining, drug discovery, and biomedical research (Luo et al., 2022).

In the financial domain, FinGPT leverages fine-tuned transformer models and parameter-efficient techniques, such as LoRA, to process real-time, multi-source financial data. By applying domain-specific NLP methods such as sentiment analysis, named entity recognition, and financial document parsing, it powers tasks like market forecasting, trading signal generation, and robo-advisory services, enabling high-precision, low-latency decision-making in finance (Yang et al., 2023).

In geosciences, OceanGPT has demonstrated the effectiveness of domain-tuned LLMs for oceanographic applications by integrating structured datasets with synthetic data generation to support marine modeling and research (Li et al., 2024).

Similarly, in the agricultural domain, IPM‑AgriGPT leverages a Generation–Evaluation Adversarial (G‑EA) framework to generate high-quality, domain-specific Q\&A pairs without manual annotation. Fine-tuned using parameter-efficient methods like LoRA, and enhanced by Agricultural Contextual Reasoning via Chain-of-Thought Distillation (ACR‑CoTD), the model captures multi-step agronomic reasoning. Trained on diverse agricultural corpora, IPM-AgriGPT supports tasks such as pest diagnosis, treatment planning, and preventive strategy generation, offering scalable, expert-level decision support in integrated pest and disease management (Gupta et al., 2024).

Building upon these advancements, we introduce \textbf{AQUA}, the first domain-specific LLM tailored for aquaculture. AQUA addresses critical industry challenges, such as disease management, feeding optimization, and hatchery operations, through the \textbf{AQUADAPT} framework, a novel agentic pipeline for high-quality data acquisition, synthetic QA generation, and expert-driven evaluation. By aligning model training with domain expertise and operational contexts, AQUA represents a significant step toward intelligent and scalable solutions in aquaculture.

\section{\textbf{Data Collection and Preprocessing}}

To build a robust aquaculture-specific language model, we curated a domain-specific corpus consisting of \textbf{55,105} (Tab. 1) documents from diverse open-access sources. Our data collection methodology incorporated multiple strategies, including web scraping, PDF extraction, and textual mining from different data sources. This approach ensured broad coverage of the aquaculture domain. 

\renewcommand{\thetable}{\arabic{table}}

\begin{table}[htbp]
\centering
\begin{tabularx}{\linewidth}{@{}X r@{}}
\toprule
\textbf{Data Source} & \textbf{Quantity} \\
\midrule
Web & 43,816 \\
Open-access Source & 11,289 \\
\bottomrule
\end{tabularx}
\caption{Data Sources and Quantities}
\label{table:data_sources}
\end{table}

To manage this large-scale ingestion and refinement process, we employed a \textbf{Data Agent}, a modular agentic component of the AQUADAPT framework. The data Agent orchestrates the entire data preprocessing pipeline, ensuring consistent formatting and high relevance across the collected corpus. For document parsing and conversion, we used two efficient Python-based tools: pymupdf4llm for robust PDF extraction and Dockling for HTML and web-text conversion (Auer et al. (2024)) (Adhikari et al. (2025)).

Once the documents were converted to plain text, the Data Agent applied a rule-based cleaning pipeline using regular expressions to remove non-informative elements such as figures, tables, headers, footers, page numbers, URLs, and citation sections. In addition, the agent standardizes the text by eliminating redundant line breaks, spacing artifacts, and special characters. This automated and scalable preprocessing ensured that the resulting corpus met the structural and semantic expectations of downstream modeling.

Thus, the refined dataset served as a clean, domain-aligned foundation for expert-guided instruction design and synthetic data generation in the later stages of the AQUADAPT pipeline.

\vspace{1em}
\noindent\textbf{Aquaculture-Specific Instruction Synthesis}
\vspace{0.5em}

Given the multidimensional nature of aquaculture science, a high-fidelity instruction set must reflect its numerous subdomains. To capture this complexity, we collaborated closely with aquaculture researchers to define formal taxonomy for instruction synthesis. Through iterative domain workshops, the dataset was structured into 11 major categories and over 60 expert-curated subcategories.

Based on expert input, we finalized a robust instruction schema consisting of seed question-answer pairs and system prompts that were representative of each subdomain. These handcrafted QA templates serve as few-shot examples to condition the language models in the later synthetic generation steps. Although the Data Agent does not participate in instruction generation, it plays a critical upstream role by ensuring that only high-quality, noise-free text is passed to other components within AQUADAPT. The structured taxonomy enables modular and parallel instruction generation, is adaptable to specific subfields of aquaculture, and supports the development of instruction-tuned models, such as AQUA, with high contextual and operational relevance.

\vspace{1em}
\noindent{The defined categories are:}
\vspace{0.5em}

\begin{itemize}[itemsep=1em]
    \item \textbf{Production Systems and Infrastructure:} Includes different ways of raising fish, like ponds, tanks, cages, recirculating systems (RAS), aquaponics, mariculture, and longlines. This helps farmers to plan and run their operations effectively.

    \item \textbf{Genetics, Breeding, and Biotechnology:} Covers breeding better fish through methods like selective breeding, genetic engineering, hybridization, CRISPR gene editing, freezing genetic material, fertilization techniques, and advanced genetic testing. These methods improve fish health, growth, and disease resistance.

    \item \textbf{Larval and Hatchery Management:} Involves designing hatcheries, encouraging fish spawning, managing baby fish nutrition, growing tiny live food (like rotifers and \textit{Artemia}), setting up nursery systems, safely transporting young fish, maintaining hatchery safety, and properly incubating eggs. This ensures that young fish survive and thrive.

\begin{figure}[h]
  \centering
  \includegraphics[width=0.4\textwidth, height=0.3\textheight]{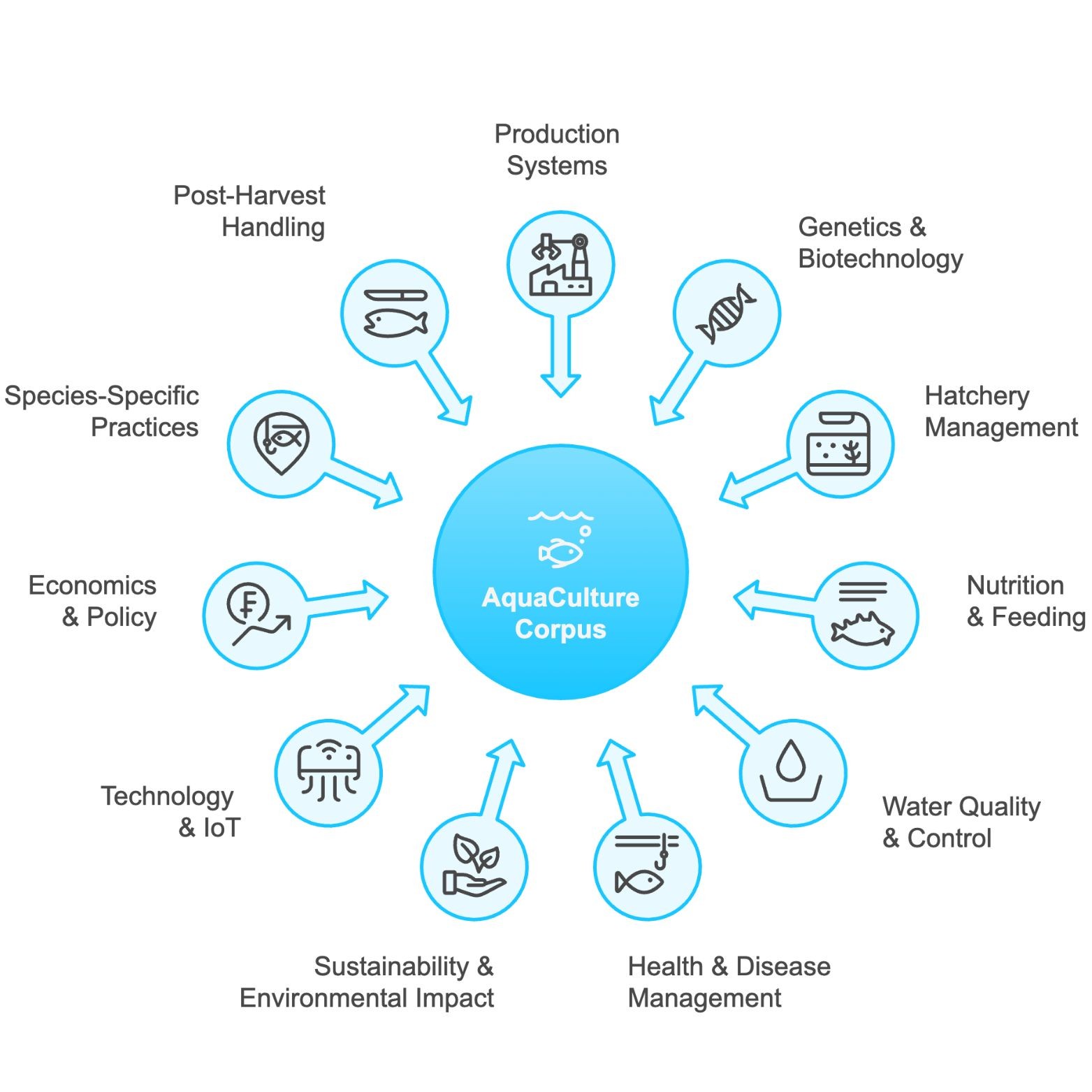}
  \caption{Aquaculture Categories}
  \label{fig:AquacultureCategories}  
\end{figure}
\vspace{-2em}
    \item \textbf{Nutrition, Feeding, and Feed Technology:} Focuses on what fish need to eat, how to make good feed, balancing protein and energy, feed efficiency (FCR), using probiotics and additives, alternative proteins, live food sources, and modern feed-making methods. Good feeding practices can improve growth and sustainability.

    \item \textbf{Water Quality and Environmental Control:} Includes managing water temperature, oxygen levels, pH balance, ammonia, nitrite, salinity, using aeration systems, biofilters, UV and ozone treatments, and removing waste effectively. Good water conditions are essential for maintaining fish health.

    \item \textbf{Health and Disease Management:} Covers identifying diseases, vaccination programs, safe use of antibiotics, managing parasites, bacteria, viruses, fungi, tracking diseases, handling disease outbreaks, using probiotics, and boosting fish immunity. This keeps fish healthy and secure.

    \item \textbf{Sustainability, Ecology, and Environmental Impact:} Involves managing waste, assessing environmental impacts, determining carrying capacity, using sustainable feed, conserving mangroves, protecting biodiversity, eco-labeling, adapting to climate change, reducing carbon footprint, and comparing wild and farmed fish practices. This promotes eco-friendly aquaculture practices.

    \item \textbf{Technology, Innovation, and IoT Applications:} Includes using smart sensors, AI monitoring, automated feeding systems, drones, blockchain for traceability, remote sensing, augmented reality training, GIS for selecting sites, machine learning for disease prediction, biofloc and nanobubble technologies, and integrated farm management systems. These innovations have improved aquaculture efficiency.

    \item \textbf{Economics, Policy, Marketing, and Governance:} Covers understanding market trends, business planning, profit analysis, following regulations, export rules, licensing, investing wisely, evaluating socio-economic impacts (such as women's roles and rural development), conducting cost-benefit studies, insurance, traceability, and certification. This helps make informed business and policy decisions possible.

    \item \textbf{Species-Specific Culture Practices:} Provides specific methods for raising popular aquaculture species like tilapia, catfish, carp, salmon, shrimp, crabs, lobsters, mussels, oysters, trout, sea bass, milkfish, grouper, snapper, tuna, eel, mullet, barramundi, flounder, halibut, and cod. This will help farmers manage each species effectively.
\vspace{2em}
    \item \textbf{Post-Harvest Handling, Processing, and Food Safety:} Covers proper harvesting, transportation, initial processing, maintaining a cold supply chain, preserving quality, grading products, packaging innovations, controlling contamination, implementing safety standards (HACCP), and obtaining food safety certifications. This ensures safe and high-quality products reach consumers.
\end{itemize}
 
\section{\textbf{A Hybrid Agent–Expert Framework for Aquaculture Instruction Data Generation}}

To ensure the high fidelity of domain-specific instruction data, we introduce an \textbf{Expert Agent}, a human-in-the-loop agent within the AQUADAPT framework responsible for seed dataset construction, quality control, and iterative refinement. This agent leverages domain expertise from aquaculture specialists to guide and validate the generation of high-quality question–answer (QA) pairs, ensuring domain alignment, semantic precision, and instructional completeness.

\subsection{\textbf{Expert-Guided Category Structuring}}
As discussed earlier in the \textbf{Aquaculture-Specific Instruction Synthesis} section, the curation process begins with each aquaculture subdomain, where experts draft representative seed question–answer pairs and system prompts that reflect realistic scenarios and challenges commonly encountered in the field.

These handcrafted examples were then used as input to the generative model (e.g., GPT-4.1), which produced candidate instruction–output pairs using a few-shot prompting approach.

\subsection{\textbf{Synthetic QA Generation and Scoring Workflow}}

Each generated QA pair undergoes expert evaluation using a 4-point Likert scale (2 to 5 stars), based on criteria such as domain relevance, factual accuracy, clarity, and completeness (see Fig. 2). Responses that received scores below 4 were flagged for revision. Experts can refine the prompt or update the few-shot examples to enhance the generation quality. This loop continues until all the QA pairs meet the required threshold.

\begin{figure}[h]
  \centering
  \includegraphics[width=0.45\textwidth, height=0.18\textheight]{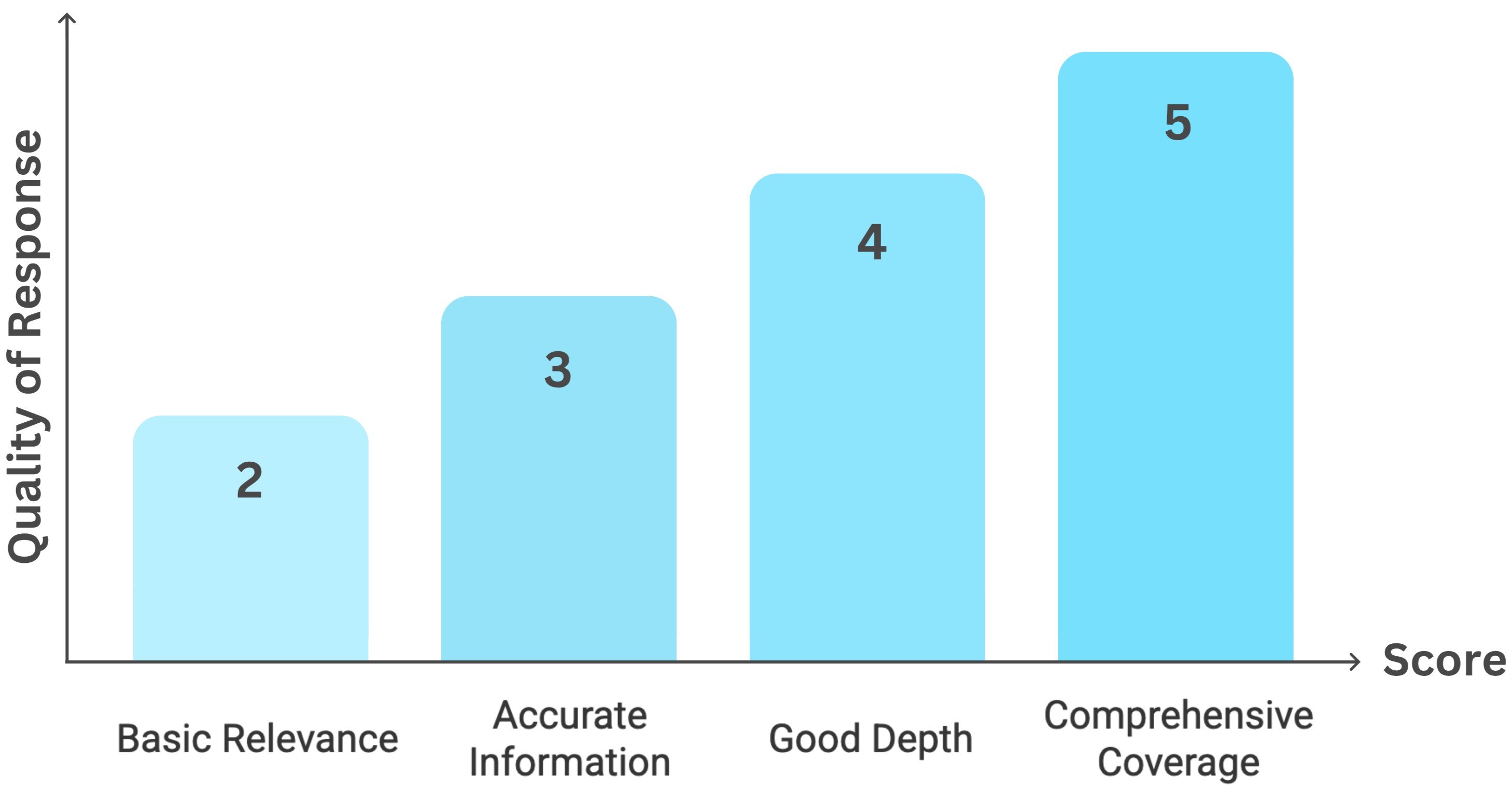}
  \caption{Aquaculture response quality ranges from basic to comprehensive.}
  \label{fig:RatingCriteria}  
\end{figure}

\subsection{\textbf{Formal Algorithmic Definition}}

Let:
\begin{itemize}[itemsep=1em]
    \item $\mathcal{C} = \{c_1, c_2, \dots, c_n\}$: Set of aquaculture categories
    \item $P_{c_i}$: Prompt template for category $c_i$
    \item $Q^{\text{seed}}_{c_i}$: Expert-authored QA seeds for $c_i$
    \item $\mathcal{G}$: Generative language model (e.g., GPT-4.1)
    \item $T$: Minimum acceptable score threshold, $T = 4$
\end{itemize}
\vspace{1em}
The \textbf{Expert Agent} operates as follows:

\vspace{1em} 

\noindent\textbf{Algorithm: Expert Agent Instruction Curation}
\vspace{0.5em}

\begin{enumerate}
    \item For each $c_i \in \mathcal{C}$:
    \begin{itemize}[itemsep=1em]
        \item Experts define $P_{c_i}$ and $Q^{\text{seed}}_{c_i}$
        \item Generate $Q^{\text{gen}}_{c_i} = \mathcal{G}(P_{c_i}, Q^{\text{seed}}_{c_i})$
    \end{itemize}
    \vspace{0.5em}
    \item For each $qa_j \in Q^{\text{gen}}_{c_i}$:
    \begin{itemize}[itemsep=1em]
        \item Compute expert score $S(qa_j) \in \{2, 3, 4, 5\}$
        \item If $S(qa_j) < T$, refine $P_{c_i}$ or $Q^{\text{seed}}_{c_i}$, regenerate $qa'_j$
    \end{itemize}
    \vspace{0.5em}
    \item Aggregate final dataset:
    \[
    D_{QA} = Q^* \cup \text{Refine}(Q^\times)
    \]
    Where:
\begin{itemize}[itemsep=1em]
    \item $Q^* = \{qa \in Q \mid S(qa) \geq T\}$
    \item $Q^\times = Q/  Q^*$
    \item $\text{Refine}(Q^\times) = \{\mathcal{G}(P', Q') \mid \text{revised by expert}\}$
\end{itemize}
\end{enumerate}

\subsection{\textbf{Output Integration}}

The resulting validated QA dataset $D_{QA}$ serves as the backbone for instruction tuning in models such as GPT-4.1 and Gemini 2.0 Flash. The Expert Agent ensures that every sample in the instruction set is not only syntactically well-formed but also semantically accurate and pedagogically aligned with real-world aquaculture needs.

This curated dataset is further used in downstream components such as QA Agent, enabling automated instruction synthesis at scale while preserving domain fidelity.

\section{\textbf{Dual-Path QA Synthesis : Literature Mining and Expert-Tuned LLMs}}

The \textbf{QA Agent} component within the \textbf{AQUADAPT} agentic framework facilitates the construction of high-quality, aquaculture-specific question--answer (QA) datasets through a dual-synthesis strategy:
\vspace{0.5em}
\begin{itemize}
    \item[(1)] Generation using a fine-tuned \textit{GPT-4.1} for instruction expansion, and
    \item[(2)] Extraction of QA pairs from scientific literature using a fine-tuned \textit{Gemini 2.0 Flash} model.
\end{itemize}

\subsection{\textbf{Expert-Guided Fine-Tuning}}

Both \textit{GPT-4.1} and \textit{Gemini 2.0 Flash} were fine-tuned on a high-quality dataset curated by the \textbf{Expert Agent}. This dataset was developed through an expert-in-the-loop process, wherein aquaculture professionals defined categories, created seed QA pairs, and iteratively refined the generated outputs based on a structured rating framework. The finalized, expert-validated QA corpus ensured that both models were imbued with accurate domain knowledge and instructional consistency.

\subsection{\textbf{BM25 Filtering for Literature Extraction}}
The QA Agent extracts QA pairs from the cleaned literature processed by the Data Agent. Given the heterogeneity of the aquaculture corpus, we applied the BM25(1) algorithm to rank and filter relevant documents. Only highly relevant content determined by domain-specific keyword scores was passed to the QA generation phase.

\begin{equation}
\scriptsize
\hspace{-0.5em}\text{BM25}(d_i, Q_{\text{aqua}})=\hspace{-1em}\sum_{q_j \in Q_{\text{aqua}}}\hspace{-1em} \text{IDF}(q_j)\cdot
\frac{f(q_j, d_i) \cdot (k_1 + 1)}{f(q_j, d_i) + k_1 \cdot \left(1 - b + b \cdot \frac{|d_i|}{\text{avgdl}}\right)}
\end{equation}

\noindent
\textbf{Where:}
\begin{itemize}[itemsep=0.5em]
    \item $f(q_j, d_i)$ is the frequency of query term $q_j$ in document $d_i$
    \item $|d_i|$ is the length of document $d_i$
    \item $\text{avgdl}$ is the average document length in the corpus
    \item $k_1$ and $b$ are hyperparameters (set to standard values: 1.5 and 0.75)
\end{itemize}

\noindent
Documents with scores above the tuned threshold $\tau$ form the filtered subset as follows:
\[
D^{\text{rel}} = \left\{ d_i \in D \mid \text{BM25}(d_i, Q_{\text{aqua}}) \geq \tau \right\}
\]

\noindent

The filtered corpus \textbf{$D^{\text{rel}}$} is then used as input for the fine-tuned \textbf{Gemini 2.0 Flash} model, which constructs QA pairs that preserve the semantic and technical integrity of the source material.

\subsection{\textbf{Gemini 2.0 Flash: QA Generation from Long-Context Documents}}

\noindent\textbf{Gemini 2.0 Flash} is selected for its capability to:
\begin{itemize}[itemsep=0.5em]
    \item Handle long-context inputs (up to \textbf{1 million tokens}),
    \item Maintain efficiency in inference,
    \item Generalize effectively from the \textit{\textbf{Expert Agent}}-tuned dataset.
\end{itemize}

These strengths enable it to generate coherent and domain-aligned QA pairs from dense scientific documents, significantly enhancing the coverage and authenticity of instruction datasets.

\subsection{\textbf{Post-Processing via \textit{Cleanup Agent}}}

The QA pairs generated from literature are subsequently passed to the \textit{\textbf{Cleanup Agent}}, which enforces rule-based quality control. This agent utilizes \textbf{Gemini 2.5 Pro} to assist in filtering out irrelevant, redundant, or semantically weak QA pairs based on syntactic structure, domain relevance, duplication, and logical completeness.

The rule set includes checks for hallucinations, off-topic drift, incomplete answers, and generic phrasings. This filtering pipeline ensures that only the most precise and contextually aligned QA pairs are retained for final inclusion in the dataset \textbf{\( \mathcal{D}_{\text{QA}}^{\text{literature-clean}} \)} significantly improves the quality and coherence of literature-derived samples.

\subsection{\textbf{Final Dataset Assembly}}

QA Agent outputs a merged and high-fidelity dataset:
\[
\mathcal{D}^{\text{final}}_{\text{QA}} = \mathcal{D}^{\text{expert}}_{\text{QA}} \cup \mathcal{D}^{\text{literature-clean}}_{\text{QA}}
\]

\noindent
Where:
\begin{itemize}[itemsep=0.5em]
  \item $\mathcal{D}^{\text{expert}}_{\text{QA}}$ represents the expert-generated synthetic QA data via GPT-4.1.
  \item $\mathcal{D}^{\text{literature-clean}}_{\text{QA}}$ represents literature-derived QA data generated by Gemini 2.0 Flash and refined by Cleanup Agent.
\end{itemize}

\vspace{1em}

This combined dataset forms the foundation for instruction-tuning \textbf{AQUA}, enabling the model to perform with both domain accuracy and contextual fluency across aquaculture-specific tasks.

\begin{figure*}[htbp]
  \centering
  \includegraphics[width=\textwidth]{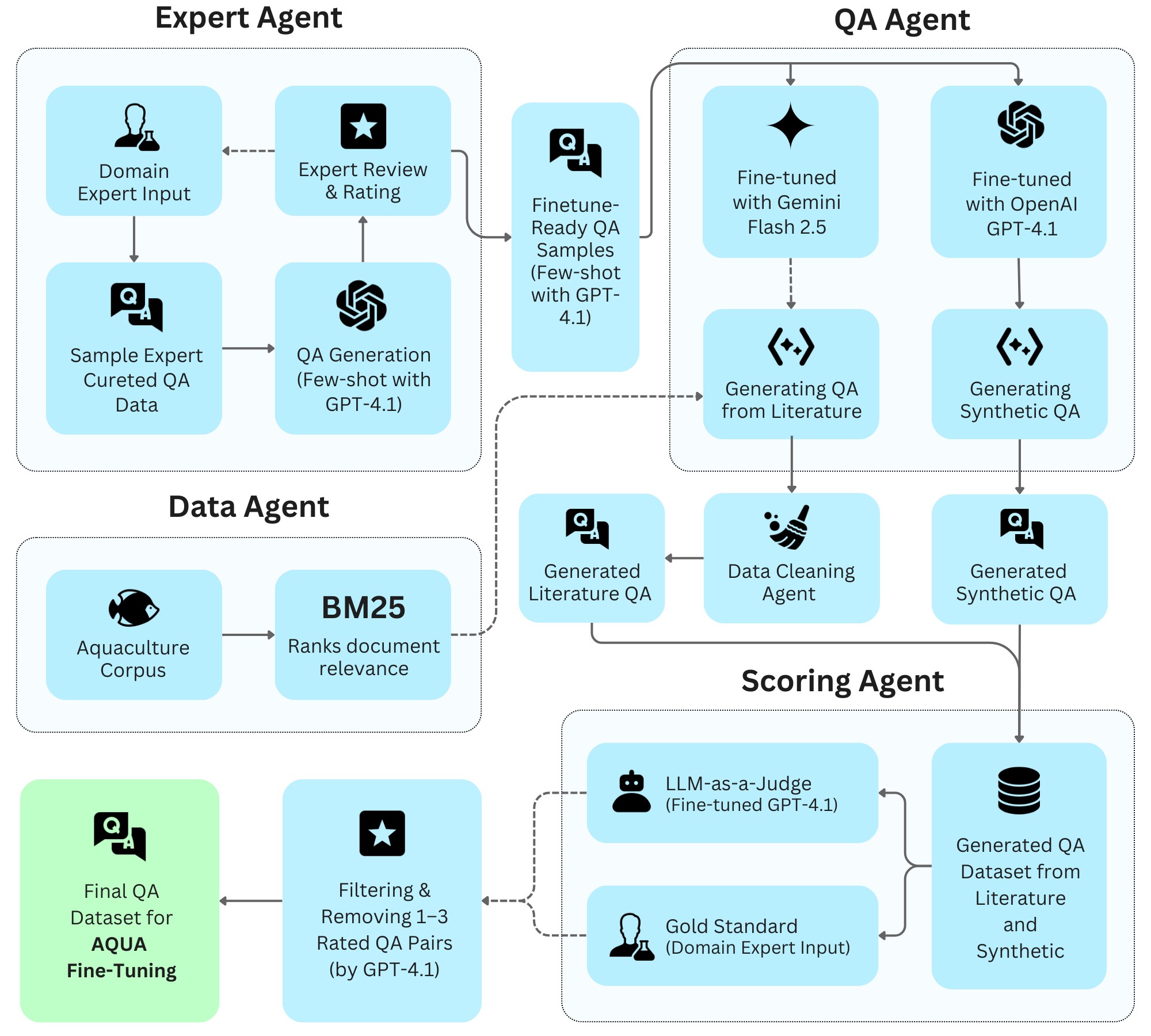}
  \caption{AQUADAPT Framework}
  \label{fig:AQUADAPTFramework}  
\end{figure*}

\section{\textbf{LLM-as-a-Judge}}
\subsection{\textbf{Establishing the Expert-Rated Gold Standard}}
The \textbf{Scoring Agent} module in our AQUADAPT framework is fundamentally anchored in an expert-rated gold standard dataset, which ensures that all automated quality assessments are grounded in authentic domain reasoning.

To build this, we randomly sampled approximately 10,000 QA pairs from the merged corpus. Each of these ~10,000 samples was meticulously evaluated by aquaculture domain experts using a structured 4-point scale (2 to 5) as done in Expert Agent. This expert-rated dataset was critical for systematically benchmarking candidate judge models within the Scoring Agent, embedding precise domain expectations directly into the evaluation pipeline.

\subsection{\textbf{Metric-Driven Evaluation to Select the Optimal LLM-as-a-Judge within Scoring Agent}}
\subsubsection{\textbf{Comparative Testing on the Gold Standard}}

We tasked \textbf{Scoring Agent} with running three advanced language models over the entire gold standard to determine which best emulated expert evaluations:

\begin{itemize}
\item \textbf{General GPT-4.1:} A zero-shot transformer trained on broad internet data.
\item \textbf{Gemini 2.5 Pro:} A cutting-edge multimodal transformer with extensive web-scale training.

\item \textbf{Fine-tuned GPT-4.1}: Specialized via expert-designed aquaculture QA data in our Expert Agent stage.
\end{itemize}
Each was prompted with few-shot examples seeded from the gold standard and tasked with producing ratings, after which we computed a comprehensive suite of metrics.

\begin{table*}[t]
\centering
\small
\begin{tabular}{|p{2.6cm}|l|c|c|c|}
\hline
\textbf{Metric Type} \rule{0pt}{2.8ex} &
\textbf{Metric} \rule{0pt}{2.8ex} &
\textbf{General GPT-4.1} \rule{0pt}{2.8ex} &
\textbf{Gemini 2.5 Pro} \rule{0pt}{2.8ex} &
\textbf{Fine-tuned GPT-4.1} \rule{0pt}{2.8ex} \\
\hline
\multirow{6}{=}{Agreement}
& Spearman’s $\rho$ (rank)     & 0.72  & 0.68  & \textbf{0.85} \\
& Kendall’s $\tau$ (ordinal)   & 0.63  & 0.59  & \textbf{0.79} \\
& Pearson correlation (linear) & 0.81  & 0.74  & \textbf{0.89} \\
& Exact match rate             & 48.3\% & 45.7\% & \textbf{63.1\%} \\
& Off-by-1 match rate          & 85.2\% & 82.9\% & \textbf{91.7\%} \\
& Mean Absolute Error (MAE)    & 0.68  & 0.73  & \textbf{0.42} \\
\hline
\multirow{2}{=}{Reliability}
& Pairwise consistency         & 81.6\% & 79.3\% & \textbf{88.5\%} \\
& Weighted Cohen’s $\kappa$    & 0.63  & 0.58  & \textbf{0.76} \\
\hline
\multirow{2}{=}{Calibration}
& Mean score (vs expert 4.06)  & 4.14  & 4.09  & \textbf{4.18} \\
& Std dev (vs expert 0.71)     & 0.67  & 0.63  & \textbf{0.66} \\
\hline
Regression
& Slope vs expert scale        & $\sim$0.89 & $\sim$0.87 & \textbf{$\sim$0.93} \\
\hline
\end{tabular}

\centering
\textbf{Tab.~2}\\
Quantitative Results from Scoring Agent Benchmarking. Only the fine-tuned GPT-4.1 consistently replicated expert ordinal preferences, numeric scales, and pairwise comparative logic.
\label{tab:scoring agent}
\end{table*}

\subsection{\textbf{Fine-Tuned GPT-4.1 as the LLM-as-a-Judge in Scoring Agent}}

Given this comprehensive metric-driven evidence, \textit{Scoring Agent} designated the \textbf{fine-tuned GPT-4.1} as the official \textit{LLM-as-a-Judge}. 

\noindent It demonstrated:

\begin{itemize}
    \item The strongest rank and linear correlations with expert scores (\( \rho = 0.85 \), \( \tau = 0.79 \), \( r = 0.89 \)).
    
    \item The lowest MAE (\( 0.42 \)) and the highest exact or near match rates, ensuring minimal average disagreement.
    
    \item The highest pairwise consistency (\(88.5\%\)) and a robust \( \kappa = 0.76 \), confirming it internalized expert comparative logic.
    
    \item A rating distribution (mean \( = 4.18 \), std \( = 0.66 \)) closely aligned with expert use of the scale.
\end{itemize}

This rigorous selection ensured that downstream automated judgments remained tightly coupled to expert quality expectations.

\subsection{\textbf{Applying Scoring Agent to Filter the Final Dataset}}

Once validated, Scoring Agent used the fine-tuned GPT-4.1 to score every QA pair in the full combined dataset. Each pair was rated on the same 2--5 scale, grounded by few-shot gold standard examples.

We then enforced a strict threshold ensuring that only QA pairs rated \textbf{4 or above} and deemed by experts to reflect \textit{good to comprehensive depth, accuracy, and clarity} were retained.

\[
\mathcal{D}^{\text{final}}_{\text{QA}} = \left\{ q_i \;\middle|\; S_i^{(L)} \geq 4 \right\}
\]
\noindent\textbf{Where:}
\begin{itemize}
    \item $\mathcal{D}^{\text{final}}_{\text{QA}}$ denotes the final curated set of high-quality question–answer pairs used to train AQUA.
    \item $q_i$ represents an individual QA pair from the combined literature and synthetic dataset.
    \item $S_i^{(L)}$ is the quality rating assigned by the \textbf{LLM-as-a-Judge} (implemented within Scoring Agent), based on few-shot prompting calibrated with expert-rated examples.
\end{itemize}

This process resulted in a final high-integrity dataset of approximately \textbf{3 million QA pairs}, each systematically validated by a scoring mechanism proven to emulate expert judgment across all critical metrics.

\subsection{\textbf{Significance of Scoring Agent’s Role}}
Through this pipeline, the Scoring Agent provides a scalable, domain-calibrated mechanism for curating instruction datasets, seamlessly bridging expert reasoning with large-scale LLM processing. By grounding automated scores in a robust expert-rated gold standard
and using multi-metric benchmarking to select the LLM-as-a-Judge, we established a rigorous methodology for ensuring data quality in AQUA, setting a template for future domain-specific LLM development.

\section{\textbf{Fine-Tuning and Evaluation of Aqua}}

\subsection{\textbf{Supervised Fine-Tuning Setup}}
We fine-tuned the open-weight Mistral-7B-Instruct-v0.3 on approximately 3 million high-quality QA pairs curated by our Scoring Agent. Each sample was validated with a minimum expert-aligned score of 4, ensuring strong factual depth and instructional clarity. This domain specialization process targeted operational, regulatory, and technical nuances of aquaculture tasks.

The model was trained for 2 epochs on 8 NVIDIA H200 GPUs over 32 hours, leveraging parameter-efficient LoRA adapters and mixed precision to handle the dataset’s scale (Hu et al., 2021)(Jiang et al., 2023)(Qwen et al., 2025) .

\begin{table}[h]
\centering
\begin{tabular}{ll}
\toprule
\textbf{Hyperparameter} & \textbf{Setting} \\
\midrule
Base model & Mistral-7B-Instruct-v0.3 \\
Fine-tuning method & LoRA \\
Optimizer & AdamW with $lr = 1\mathrm{e}{-4}$ \\
Scheduler & Cosine, 5,000 warmup steps \\
Epoch & 2 \\
Batch size & 2 \\
Gradient Accumulation & 8 \\
Effective batch size & 16 samples \\
Max sequence length & 2048 tokens \\
Precision & bf16 \\
Max gradient norm & 1.0 \\
\bottomrule
\end{tabular}

\centering
\textbf{Tab.~3}\\
Training hyperparameters for fine-tuning Mistral-7B-Instruct-v0.3
\label{tab:training-hparams}
\end{table}

\vspace{-1.5em}

\subsection{\textbf{Evaluation Metrics on Validation QA Pairs}}
The resulting AQUA model was evaluated on a held-out validation set of approximately 20,000 QA pairs \textbf{(https://huggingface.co/datasets/KurmaAI/AQUA-Test-Dataset)} using reference responses from expert-rated data. Standard NLG metrics confirmed that the model reliably internalized aquaculture-specific language and multi-step instruction pattern(Papineni et al., 2002)(Vaswani et al., 2023).

\begin{table}[htbp]
\centering
\begin{tabularx}{0.5\textwidth}{l r X}
\toprule
\textbf{Metric} & \textbf{Value} & \textbf{Interpretation} \\
\midrule
BLEU-4  & 49.19 & Strong multiword phrase fidelity. \\
ROUGE-1 & 51.45 & High coverage of key domain terms. \\
ROUGE-2 & 30.98 & Effective short technical phrase recall. \\
ROUGE-L & 45.09 & Preserved logical sequence of instructions. \\
\bottomrule
\end{tabularx}

\vspace{1pt}
\centering
\textbf{Tab.~4}\\Evaluation metrics and their interpretation for generated technical responses.
\label{tab:metric-interpretation}
\end{table}
\vspace{-1em}
\section{\textbf{Comparative Evaluation of Model Outputs Across Aquaculture Domains}}

To rigorously validate the domain specialization of our fine-tuned \textbf{AQUA (Mistral-7B)} model, we conducted a structured comparative assessment against two prominent general-purpose instruction-tuned models: \textbf{Qwen-2.5-7B} and \textbf{Llama-3.1-8B}. This evaluation was designed to benchmark performance across four critical areas of aquaculture, systematically identified through expert-driven topic mapping:

\begin{itemize}
    \item \textbf{Fish Health Management}
    \item \textbf{Feeding Strategies and Nutrition}
    \item \textbf{Aquaculture Production and Economics}
    \item \textbf{Water Quality Management}
\end{itemize}

\onecolumn

\begin{table}[H]
\centering
\section*{\textbf{Fish Health Management}}
\noindent
\textbf{Q:} What is a key management practice to reduce fish mortality in an aquaculture system?

\vspace{1em}
\renewcommand{\arraystretch}{1.3}
\small
\begin{tabularx}{\textwidth}{@{}p{2.2cm}p{8cm}p{1cm}p{4cm}@{}}
\toprule
\textbf{Model} & \textbf{Response} & \textbf{Rating} & \textbf{Rationale} \\
\midrule

Expert & “Immediately implement \textbf{regular health monitoring} by inspecting fish daily for behavioral changes (lethargy, erratic swimming) and physical signs (lesions, discoloration). Early detection enables prompt treatment such as targeted probiotics or therapeutics and can reduce mortality by over 30\%.” 
& 5 & Full procedural depth, specific metrics, diagnostics, and treatment steps. \\

Qwen-2.5 & “Regular health monitoring helps detect signs of disease or stress early observing behavior changes and physical symptoms in farmed fish enables prompt treatment and significantly lowers mortality rates.” 
& 3 & Accurate concept but lacks detail on frequency, diagnostics, and recordkeeping. \\

Llama-3.1 & “Conduct routine health checks inspecting fish for signs of illness or abnormal behavior allows farmers to intervene early and prevent large-scale losses.” 
& 2 & Basic advice without actionable steps, thresholds, or monitoring protocols. \\

AQUA & “\textbf{Begin daily health checks} observe behavior and sample gill and skin condition every 72 hours. Record all findings in a digital log and act on any anomalies immediately. This should be combined with periodic blood parameter tests and swift isolation of the affected stock. This comprehensive protocol ensures full coverage of fish health management.”
& 5 & Comprehensive coverage: precise intervals, diagnostics, record keeping, and isolation steps. \\

\bottomrule
\end{tabularx}
\end{table}

\begin{table}[H]
\centering
\section*{\textbf{Feeding Strategies \& Nutrition}}
\noindent
\textbf{Q:} On average, how much feed is required to produce a pound of gain in fish?

\vspace{1em}
\renewcommand{\arraystretch}{1.3}
\small
\begin{tabularx}{\textwidth}{@{}p{2.2cm}p{8cm}p{1cm}p{4cm}@{}}
\toprule
\textbf{Model} & \textbf{Response} & \textbf{Rating} & \textbf{Rationale} \\
\midrule

Expert & “Under optimal conditions, most finfish achieve a \textbf{feed conversion ratio (FCR)} of $\sim$1.5:1, meaning 1.5 lb of feed yields 1 lb of weight gain. Monitor FCR weekly and adjust feed formulation or feeding frequency to maintain this target and minimize waste.” 
& 5 & Includes target FCR, monitoring frequency, and adjustment strategy. \\

Qwen-2.5 & “Feed conversion ratios vary by species, but many cultured fish require around \textbf{2.5 lb of feed per pound of growth}, depending on factors like diet quality and environmental conditions.” 
& 2 & Incorrect FCR value, not aligned with industry standards. \\

Llama-3.1 & “A commonly cited average feed conversion ratio is about \textbf{2 lb of feed for each pound of growth}, though optimal systems can achieve lower ratios.” 
& 3 & General accurate range but lacks target setting and optimization guidance. \\

AQUA & “Target \textbf{1.5 lb feed per 1 lb gain} (FCR = 1.5). To optimize, record feed input and growth weekly; if FCR rises above 1.6, adjust protein content or feeding frequency. For comprehensive management, also consider temperature- and size-adjusted feeding rates to refine efficiency.” 
& 5 & Correct ratio with optimization steps. \\

\bottomrule
\end{tabularx}
\end{table}

\vspace{1em}

\begin{table}[H]
\centering
\section*{\textbf{Aquaculture Production \& Economics}}
\noindent
\textbf{Q:} How much of the world’s seafood comes from aquaculture?

\vspace{1em}
\renewcommand{\arraystretch}{1.3}
\small
\begin{tabularx}{\textwidth}{@{}p{2.2cm}p{8cm}p{1cm}p{4cm}@{}}
\toprule
\textbf{Model} & \textbf{Response} & \textbf{Rating} & \textbf{Rationale} \\
\midrule

Expert & “Aquaculture now supplies 50\% of global seafood production (FAO 2023). This share has steadily risen due to intensification, genetic improvement, and feed advancements, underscoring aquaculture’s critical role.” 
& 5 & Precise statistic with contextual drivers, trends, and significance. \\

Qwen-2.5 & “Recent estimates show that approximately \textbf{75\% of seafood consumed in some markets} is produced by aquaculture, reflecting strong growth in farmed fish and shellfish.” 
& 2 & Inaccurate global statistic, misrepresents actual production share. \\

Llama-3.1 & “According to FAO data, \textbf{aquaculture now contributes around 50\%} of global seafood supply, a figure that has climbed steadily over the past two decades.” 
& 4 & Accurate but minimal context on historical trends or drivers. \\

AQUA & “Currently \textbf{50\% of consumed seafood} is farmed (FAO 2023). This highlights aquaculture’s parity with wild fisheries. Year-on-year, production grew \textasciitilde5\%, driven by better feeds and biosecurity-key for sustained global protein supply.” 
& 4 & Adds growth rate but lacks deeper economic analysis for full expert depth. \\

\bottomrule
\end{tabularx}
\end{table}

\vspace{1em}

\begin{table}[H]
\centering
\section*{\textbf{Water Quality Management}}
\noindent
\textbf{Q:} How do bacteria affect dissolved oxygen (DO) in water bodies?

\vspace{1em}
\renewcommand{\arraystretch}{1.3}
\small
\begin{tabularx}{\textwidth}{@{}p{2.2cm}p{8cm}p{1cm}p{4cm}@{}}
\toprule
\textbf{Model} & \textbf{Response} & \textbf{Rating} & \textbf{Rationale} \\
\midrule

Expert & “During decomposition of organic matter, bacteria \textbf{consume DO}, driving levels below 3 mg/L in high-load systems. This bacterial oxygen demand is a primary factor in pond hypoxia; implement aeration or biofiltration to maintain DO above 5 mg/L.” 
& 5 & Detailed mechanism, thresholds, and mitigation strategies. \\

Qwen-2.5 & “As bacteria decompose organic material, they consume dissolved oxygen, leading to \textbf{decreased DO} levels and potentially causing hypoxic conditions in aquatic systems.” 
& 4 & Correct mechanism but omits threshold values and corrective measures. \\

Llama-3.1 & “Bacterial respiration during decomposition uses up oxygen, which \textbf{decreases dissolved oxygen} concentrations and can stress or kill aquatic animals if not managed.” 
& 4 & Accurate but lacks specific thresholds and mitigation guidance. \\

AQUA & “Bacteria \textbf{reduce DO} by consuming oxygen during breakdown of organic wastes, often causing levels to fall below 3 mg/L in unmanaged systems. To offset this, deploy continuous aeration and incorporate biofilters; monitor DO hourly and maintain above 5 mg/L for optimal health.” 
& 5 & Combines mechanism, threshold values, monitoring frequency, and corrective actions. \\

\bottomrule
\end{tabularx}
\end{table}
\vspace{1em}

\twocolumn

\begin{figure*}[htbp]
  \centering
  \includegraphics[width=\textwidth]{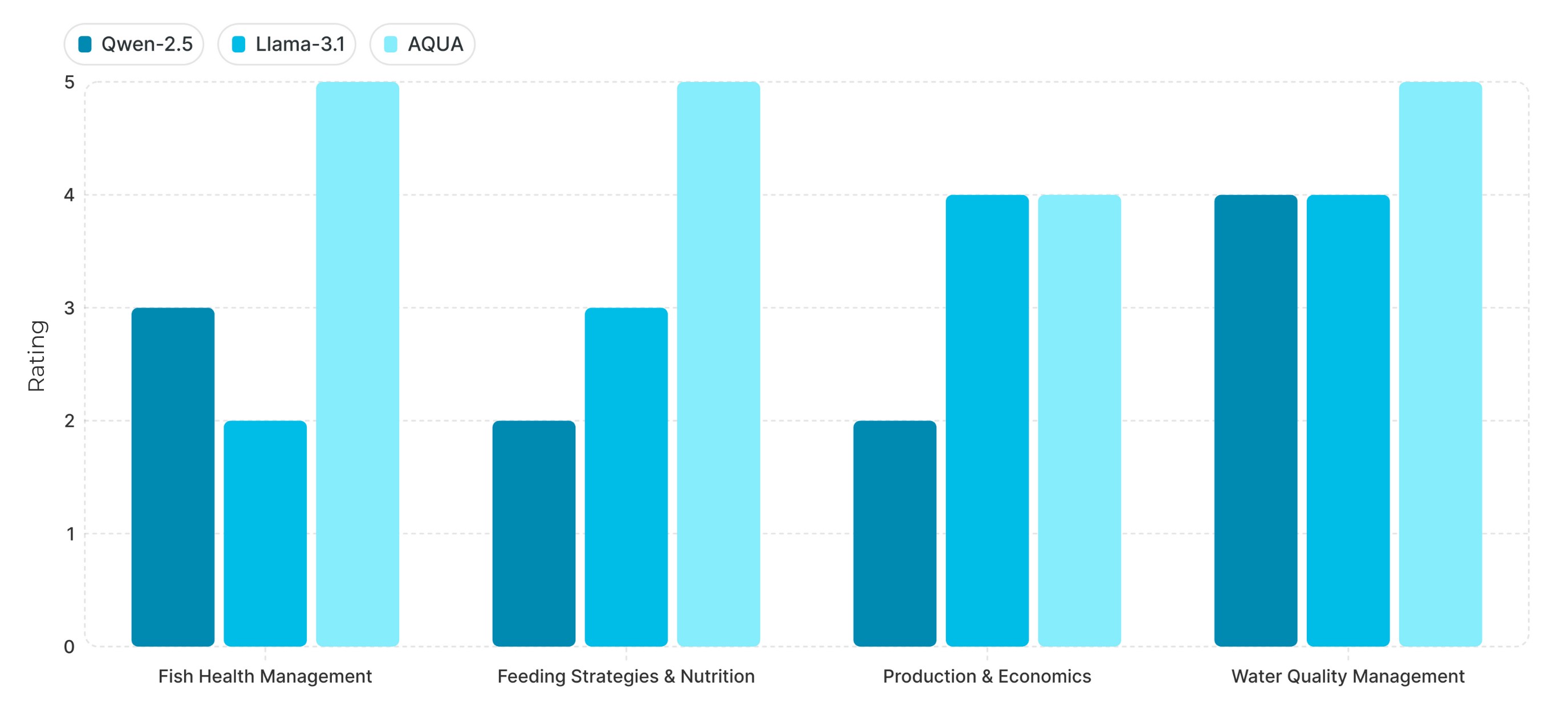}  
  \caption{AQUA consistently achieves higher ratings, with an average of 4.75, followed by Llama at 3.25 and Qwen at 2.75.}
  \vspace{1em}
  \textbf{Try AQUA-7B at Huggingface:} \url{https://huggingface.co/KurmaAI/AQUA-7B}
  \label{fig:ComparativeResults}  
\end{figure*}

Each topic was represented by a carefully constructed question grounded in realistic operational contexts, reflecting common decision points and technical challenges encountered by aquaculture practitioners. The model responses were evaluated using a domain-aligned rubric that assessed:

\begin{enumerate}
    \item Factual precision and adherence to aquaculture standards,
    \item Depth of contextual explanation and coverage of causative processes,
    \item Use of quantitative thresholds and industry-specific benchmarks, and
    \item Clarity and actionability of guidance provided.
\end{enumerate}

Ratings were assigned based on how well each response met these criteria relative to an expert’s reference standard.

\section{\textbf{Comparative Results and Domain Advantage}}

The comparative study clearly demonstrated that \textbf{AQUA}, trained on a rigorously curated, expert-rated aquaculture dataset, consistently outperformed general-purpose instruction models across all four topical areas. Specifically:

\subsubsection{\textbf{Fish Health Management}}
\noindent
\textbf{AQUA} delivered detailed procedural guidance, including monitoring intervals, sampling protocols, and isolation measures, aligning closely with expert practices for proactive disease prevention.

\subsubsection{\textbf{Feeding Strategies and Nutrition}}
\noindent
Unlike \textbf{Qwen} and \textbf{Llama}, which often provided generalized or incorrect feed conversion ratios (FCR), \textbf{AQUA} reliably cited optimal FCR values and offered clear tactics for sustaining feed efficiency.

\subsubsection{\textbf{Aquaculture Production and Economics}}
\noindent
\textbf{AQUA} not only reported accurate global production shares but contextualized these statistics with multi-year trends and underlying productivity drivers, offering insights that extended beyond superficial figures.

\subsubsection{\textbf{Water Quality Management}}
\noindent
\textbf{AQUA}'s responses integrated precise dissolved oxygen thresholds with concrete mitigation strategies, addressing both immediate remediation and long-term system stability.

These results underscore the critical advantages of domain-specific fine-tuning. While generalist models can approximate factual content, they frequently lack the procedural granularity, context-dependent recommendations, and operational benchmarks necessary for high-stakes decision-making in specialized industries such as aquaculture. In contrast, \textbf{AQUA} effectively bridges this gap, producing outputs that mirror expert-level reasoning and actionable guidance that are essential for the aquaculture sector.

\section{\textbf{SLM Integration in IoT Systems for Real-Time Water Quality Management}}

As a forward-looking extension of this work, we explored the integration of specialized small language models (SLMs) into IoT-based aquaculture infrastructure to enable real-time edge-level water quality monitoring and advisories. Drawing on the architecture illustrated in related work (see extracted reference on AQUA-1B with Raspberry Pi and MQTT), such systems present a promising pathway for deploying lightweight LLMs directly on farms.

\subsection{\textbf{System Overview}}

The architecture integrates:
\begin{itemize}
    \item Low-cost sensor arrays to measure critical parameters:
    \begin{itemize}
        \item Dissolved Oxygen (DO), Temperature, pH, Ammonia, and Turbidity.
    \end{itemize}
    \item Raspberry Pi as an edge compute unit, interfacing analog sensors via MCP3008 ADC and digital sensors like DS18B20 for temperature through the 1-Wire protocol.
    \item MQTT protocol for lightweight, asynchronous communication, publishing sensor data to local or cloud brokers under structured topics (e.g., \texttt{/aqua/ammonia}).
\end{itemize}

\subsection{\textbf{Role of Small Language Models (SLMs)}}

A small but domain-specialized model (e.g., \textbf{AQUA-1B}, a 1-billion parameter LLM fine-tuned on aquaculture datasets) can:
\begin{itemize}
    \item Run locally on Raspberry Pi hardware, minimizing latency and ensuring functionality even without stable internet.
    \item Operate via a cloud inference API, processing structured sensor prompts to provide on-demand analytics.
\end{itemize}

\subsection{\textbf{Real-Time Decision Support}}

The system continuously compiles sensor data into prompts for the SLM to:
\begin{itemize}
    \item Generate risk analyses (e.g., identifying likely causes for low DO given simultaneous ammonia spikes).
    \item Recommend immediate corrective actions, such as adjusting aeration or water exchange rates.
\end{itemize}

\subsection{\textbf{User Interaction and Alerting}}

\begin{itemize}
    \item Integrates with Telegram bots to automatically notify operators when critical thresholds are exceeded (e.g., DO $<$ 3 mg/L or ammonia $>$ 0.5 ppm).
    \item Supports direct natural language queries from farmers (e.g., ``Why is my pond showing high turbidity today?''), with instant, domain-tuned responses.
\end{itemize}

This integration of lightweight language models into real-time monitoring systems represents a critical step toward fully autonomous, intelligent aquaculture operations, combining IoT hardware, domain expertise, and scalable AI in a single cohesive platform.

\section{\textbf{Illustrative Use Case: Real-Time SLM-Driven Advisory via IoT for Water Quality Management}}

To demonstrate the practical integration of our approach in an operational aquaculture setting, we present an example scenario where a lightweight small language model (SLM), deployed on an IoT-based monitoring system, autonomously interprets sensor data and communicates advisories to the farmer through a Telegram interface.

\subsection{\textbf{Scenario Overview}}

The sensor readings captured by the IoT system were as follows:

\begin{table}[h!]
\centering
\renewcommand{\arraystretch}{1.4}
\begin{tabularx}{\linewidth}{|p{2.5cm}|p{1.6cm}|X|}
\hline
\textbf{Parameter} & \textbf{Value} & \textbf{Interpretation} \\
\hline
Dissolved Oxygen (DO) & 4.87 mg/L & Within acceptable range \\
\hline
Temperature & 29.65°C & Elevated, warning level \\
\hline
pH & 7.58 & Mild deviation, warning level \\
\hline
Ammonia & 14.09 mg/L & Critically high \\
\hline
Turbidity & 6.84 NTU & Elevated, warning level \\
\hline
\end{tabularx}
\centering
\textbf{Tab.~5 }\\Sensor Readings and Real-Time Interpretations.
\label{tab:readings}
\end{table}

\subsubsection*{\textbf{Automated Analysis by the SLM}}

The on-device SLM processed these readings and contextualized the multiparameter data stream. Unlike traditional threshold alarms that may operate on isolated triggers, the SLM combined elevated ammonia, increased temperature, and turbidity levels to recognize a compound-stress scenario. A structured advisory message prioritizing interventions based on urgency and operational feasibility was generated.

\vspace{0.5em}
\noindent\textbf{AQUA IoT Alert – Pond Health Summary}
\vspace{0.5em}

\noindent\textbf{Current Readings:}
\begin{itemize}
    \item \textbf{Dissolved Oxygen (DO)}: 4.87 mg/L \hfill (Good)
    \item \textbf{Temperature}: 29.65°C \hfill (High)
    \item \textbf{pH}: 7.58 \hfill (Mildly elevated)
    \item \textbf{Ammonia}: 14.09 mg/L \hfill (Critical)
    \item \textbf{Turbidity}: 6.84 NTU \hfill (High)
\end{itemize}

\noindent\textbf{Interpretation:}

Ammonia levels are critically elevated ($>$14 mg/L), posing an immediate risk of gill damage and suppressed immunity. Elevated temperature intensifies ammonia toxicity, whereas increased turbidity indicates an excessive organic load.

\noindent\textbf{Recommended Immediate Actions:}

\begin{enumerate}
    \item Activate or increase aeration to improve oxygen exchange and facilitate ammonia stripping.
    \item Initiate a partial water exchange (20–30\%) with cooler, clean water to dilute ammonia.
    \item Temporarily reduce feeding to limit further ammonia accumulation.
    \item Reassess water quality in 6–8 hours to monitor improvement.
    \item Inspect aerators, pumps, and filters to ensure optimal functioning.
\end{enumerate}

\textbf{Note:} Continuous monitoring will proceed every 15 minutes; you will receive additional alerts if parameters deteriorate.

\section{\textbf{Limitations}}

Despite the demonstrated strengths of AQUA, particularly in delivering detailed, operationally sound aquaculture guidance, several limitations arise from the very nature of large-scale, domain-specific fine-tuning. These were categorized as follows:

\subsection*{\textbf{A. Domain Generalization vs. Over-Specificity}}

Although AQUA was trained on an extensive, high-quality aquaculture dataset encompassing a broad global perspective, it sometimes exhibits over-contextualization even in general queries. For example:

\begin{quote}
\textit{“How can I optimize feed management for my fish farm?”}
\end{quote}

may elicit responses framed around specific regional case studies, such as Indonesian tilapia or Norwegian salmon, without explicit regional prompts. This behavior likely stems from the high density of region-specific examples in the training corpus, leading the model to default to these as implicit standards.

\subsection*{\textbf{B. Potential Rigidity from Immense Domain Anchoring}}

Fine-tuning on a very large, richly detailed aquaculture corpus has anchored the model deeply into prevailing practices documented across scientific and industry literature. While this ensures exceptional alignment with aquaculture norms, it introduces certain trade-offs:

\begin{itemize}
    \item \textbf{Narrower interpretive bandwidth:} The model may sometimes prioritize mainstream or widely documented approaches, underrepresenting innovative, emerging, or experimental techniques that lack extensive literature backing.
    \item \textbf{Reduced conversational flexibility:} Compared to purely general-purpose LLMs, AQUA may generate answers that are more formal, technical, or prescriptive, sometimes at the expense of broader exploratory dialogue styles.
\end{itemize}

This underscores a common tension in heavy domain specialization: achieving high operational fidelity can slightly reduce the model’s capacity for creatively reasoning beyond dominant documented paradigms.

\subsection*{\textbf{C. Model Sensitivity to Underrepresented Contexts}}

Despite encompassing interdisciplinary topics such as environmental economics, sustainability assessments, and water resource governance, certain localized practices especially from underrepresented small-scale or indigenous aquaculture systems remain sparsely covered in global open-access literature. As a result, AQUA’s recommendations in such contexts may default to general best practices without sufficiently nuanced cultural or smallholder economic framing.

\section{\textbf{Future Work}}

Looking ahead, several targeted developments will further advance \textbf{AQUA} into a fully autonomous system under the broader vision of \textit{\textbf{Aquaculture Intelligence}} a next-generation intelligent platform for aquaculture management:

\subsection*{\textbf{A. Agentic Workflows for Full Automation}}

Future work will build a multi-agent architecture, where specialized agents process sensor streams, perform multiparameter diagnostics, and dynamically query AQUA’s knowledge base to generate precise, context-aware recommendations that minimize the need for continuous human intervention.

\subsection*{\textbf{B. Real-Time Visual Models}}

Integrating computer vision modules into this agentic workflow will enable automatic detection of fish health problems (e.g., lesions, abnormal pigmentation), growth estimation, and behavioral stress cues such as irregular schooling or surface gasping providing a rich, multi-modal layer of system insight.

\subsection*{\textbf{C. Regional and Context-Aware Adaptation}}

Next stages will incorporate user-specific conditioning based on location, production system type, and target species, ensuring that recommendations seamlessly adapt to local practices and regulatory standards.

\subsection*{\textbf{D. Continual Learning and Knowledge Refresh}}

Automated pipelines periodically update AQUA’s knowledge using new research publications, case studies, and aggregated farm telemetry, keeping the system aligned with evolving industry practices without sacrificing hard-won domain expertise.

\section{\textbf{Vision}}

Collectively, these advancements will establish \textit{Aquaculture Intelligence} as a comprehensive, agent-driven ecosystem that combines real-time sensory data, visual fish health analytics, and extensive aquaculture expertise to autonomously manage farm operations, maximizing productivity, animal welfare, and environmental sustainability.

\section{\textbf{Acknowledgment}}

We would like to express our sincere gratitude to the aquaculture researchers and domain experts who served as mentors and evaluators throughout the development of this work. Their contributions were instrumental in designing seed instruction sets, rating QA pairs, and providing deep domain feedback that shaped the learning trajectory of the large language model. Their role as teachers to the LLM was invaluable in aligning the model's behavior with real-world aquaculture challenges.

We also acknowledge the generous computational support provided by Nebius, who enabled large-scale model fine-tuning through credits for NVIDIA H200 GPU clusters. We thank the Microsoft for Startups program for granting access to Azure OpenAI infrastructure and GPT-4.1 training resources. Additionally, we are grateful to Google for providing API and model fine-tuning credits for Gemini through its research initiative.
The success of this work was made possible through the collaborative ecosystem of technical, academic, and cloud infrastructure support, without which the development of AQUA and the AQUADAPT framework would not have been feasible

\section{\textbf{Conclusion}}
\vspace{0em}
This work introduced AQUA, a domain-specialized large language model meticulously fine-tuned for aquaculture. Built upon a rigorously engineered agentic framework, AQUADAPT, our methodology combined automated data extraction, expert-in-the-loop validation, synthetic QA generation, and multi-stage evaluation to curate a high-fidelity dataset that captures the complexity and operational nuances of aquaculture systems. This process was essential for aligning AQUA’s outputs with real-world expert-level reasoning across diverse aquaculture subdomains.

Through extensive comparative evaluations, AQUA consistently outperformed general-purpose instruction models such as Qwen and Llama, delivering more precise, context-sensitive, and practically actionable insights. Its superior performance spanned critical areas including fish health management, feed optimization, water quality control, and economic planning. The use of multi-agent processes from initial data gathering through rigorous expert scoring and iterative refinement proved central to ensuring AQUA’s exceptional domain alignment and decision-support fidelity.

Additionally, our exploration of lightweight SLM deployments on IoT platforms demonstrated the viability of extending AQUA to edge environments, enabling real-time sensory interpretation and automated, context-aware advisories directly on farms. This lays a crucial groundwork for future closed-loop aquaculture management systems that tightly integrate environmental telemetry with intelligent recommendations.

Looking ahead, we aim to advance AQUA into a comprehensive multi-agent ecosystem under the broader vision of Aquaculture Intelligence. This encompasses dynamic regional adaptation, visual fish health diagnostics, continual learning from new scientific and operational data, and robust explainability features. Together, these innovations will drive the evolution of precision aquaculture delivering optimized productivity, improved animal welfare, and enhanced environmental stewardship across the global aquaculture landscape.

\vspace{12pt}
\end{document}